\documentclass{article}
\usepackage[utf8]{inputenc}
\usepackage[T1]{fontenc}
\usepackage{longcat_style}

\usepackage{amsmath,amsfonts,bm}

\def\eqref#1{equation~\ref{#1}}

\def\1{\bm{1}}

\DeclareMathAlphabet{\mathsfit}{\encodingdefault}{\sfdefault}{m}{sl}
\SetMathAlphabet{\mathsfit}{bold}{\encodingdefault}{\sfdefault}{bx}{n}

\usepackage{latexsym,microtype,inconsolata}
\usepackage{graphicx,booktabs,tabularx,array,multirow,enumitem,placeins,flafter,xcolor}
\usepackage{xspace,algorithm}
\usepackage[noend]{algpseudocode}
\usepackage{wrapfig,needspace}
\usepackage{subcaption}
\usepackage[most]{tcolorbox}
\usepackage{makecell,fvextra}
\usepackage[round,authoryear]{natbib}
\usepackage{url,xurl,hyperref}
\usepackage{cleveref}
\hypersetup{colorlinks=true,linkcolor=blue,urlcolor=blue,citecolor=blue,
  pdftitle={Rep2Skill: Representation-Guided Skill Self-Evolution for LLM Agents},
  pdfauthor={Kaixing Zhang et al.}}
\DeclareCaptionLabelSeparator{pipe}{\,|\,}
\renewcommand{\shorttitle}{Rep2Skill: Representation-Guided Skill Self-Evolution}
\setlist[itemize]{leftmargin=*}
\setlist[enumerate]{leftmargin=*}

\newcolumntype{Y}{>{\raggedright\arraybackslash}X}
\newcolumntype{P}[1]{>{\raggedright\arraybackslash}p{#1}}
\newcommand{\rowhead}[2]{\multicolumn{#1}{l}{\textit{#2}}\\}

\definecolor{promptheader}{RGB}{52,52,52}
\definecolor{promptborder}{RGB}{95,95,95}

\newtcolorbox{promptbox}[1]{
    enhanced,
    breakable,
    colback=white,
    colframe=promptborder,
    boxrule=0.8pt,
    arc=0pt,
    outer arc=0pt,
    left=5pt,
    right=5pt,
    top=8pt,
    bottom=8pt,
    title={\textbf{#1}},
    coltitle=white,
    colbacktitle=promptheader,
    fonttitle=\large\bfseries,
    boxed title style={
        sharp corners,
    },
    attach boxed title to top left={
        xshift=0pt,
        yshift=0pt,
    },
    boxed title size=title,
}

\newcommand{\method}{\textsc{Rep2Skill}\xspace}

\renewcommand{\headeright}{%
  \includegraphics[height=10pt]{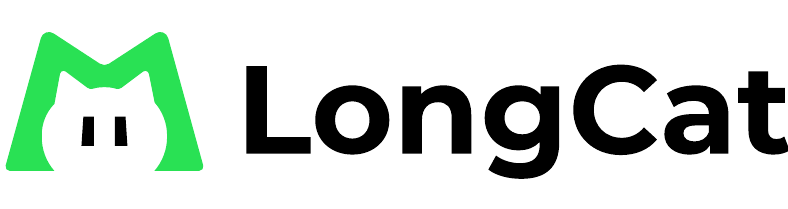}%
}

\title{Rep2Skill: Representation-Guided Skill Self-Evolution for LLM Agents}

\author{
\textbf{Kaixing Zhang}\textsuperscript{1},
\textbf{Changming Li}\textsuperscript{1,2},
\textbf{Yingdong Shi}\textsuperscript{1},
\textbf{Zheng Zhang}\textsuperscript{1,2}\\
\textbf{Kaitao Song},
\textbf{Wenjie Shi}\textsuperscript{2},
\textbf{Jingang Wang}\textsuperscript{2},
\textbf{Kan Ren}\textsuperscript{1*}
\\
\textsuperscript{1}ShanghaiTech University
\qquad
\textsuperscript{2}Meituan
\\
\small{
\textsuperscript{*}\textbf{Correspondence:}
\href{renkan@shanghaitech.edu.cn}{renkan@shanghaitech.edu.cn}
}
}

\begin{document}
\maketitle
\begin{abstract}
Textual skills enable large language model (LLM) based agents to accumulate reusable procedural knowledge without updating model parameters.
Yet existing skill evolution remains largely confined to the text space: an optimizer must diagnose success and failure patterns, and revise skills solely from long execution trajectories and sparse task outcomes.
This text-only paradigm leaves the agent's internal representations, which contain rich records of its evolving execution state, outside the skill optimization loop.
\textit{We ask whether an agent can improve its external textual skills by reflecting on its own internal representations.} 
We introduce \method, a representation-guided framework for self-evolution on agent skills.
Specifically, upon the collected agent rollouts, \method models their internal model representation trajectories to localize turns that deviate from successful execution dynamics, and it further interprets these signals alongside the execution contexts as actionable textual feedback for targeted skill revision.
Experiments on two widely-adopted agent environments with two open-source LLMs show that \method consistently outperforms text-only approaches in the self-evolution setting, where the same LLM serves as both executor and optimizer without a stronger external model.
This establishes a promising direction moving agent self-improvement beyond text-only reflection.
\end{abstract}

\section{Introduction}
\label{sec:intro}Large Language Model (LLM) agents have increasingly demonstrated strong performance on multi-step, long-horizon tasks~\citep{yao2023react,wang2023voyager,shridhar2021alfworld,merrill2026terminalbench}. 
Skills, typically composed of reusable textual instructions, procedural rules, and task-solving strategies, provide agents with explicit guidance for accomplishing complex tasks~\citep{anthropic2025agentskills,zhou2026comprehensive}.
Early agent skills were largely handcrafted by humans. 
To obtain higher quality skills at scale, recent work has explored skill evolution, where language model agents iteratively refine reusable textual skills from execution experience while keeping their model parameters fixed~\citep{alzubi2026evoskill,ni2026trace2skill,yang2026skillopt,tang2026wikiskill}.

Existing skill evolution methods predominantly rely on explicit textual artifacts, such as trajectory text~\citep{ni2026trace2skill,yang2026skillopt} and task-level outcomes~\citep{alzubi2026evoskill}.
Despite their different optimization strategies, these approaches largely require the optimizer to infer from observable artifacts which parts of an execution contributed to failure, why they failed, and how the skill should be revised accordingly. 
This dependence places substantial burden on the LLM-based skill optimizer to attribute failures, diagnose their causes, and generate appropriate skill updates, especially when the optimizer model's capability is limited.

\begin{wrapfigure}{R}{0.45\textwidth}
    \centering\captionsetup{font=footnotesize}
    \includegraphics[width=\linewidth]{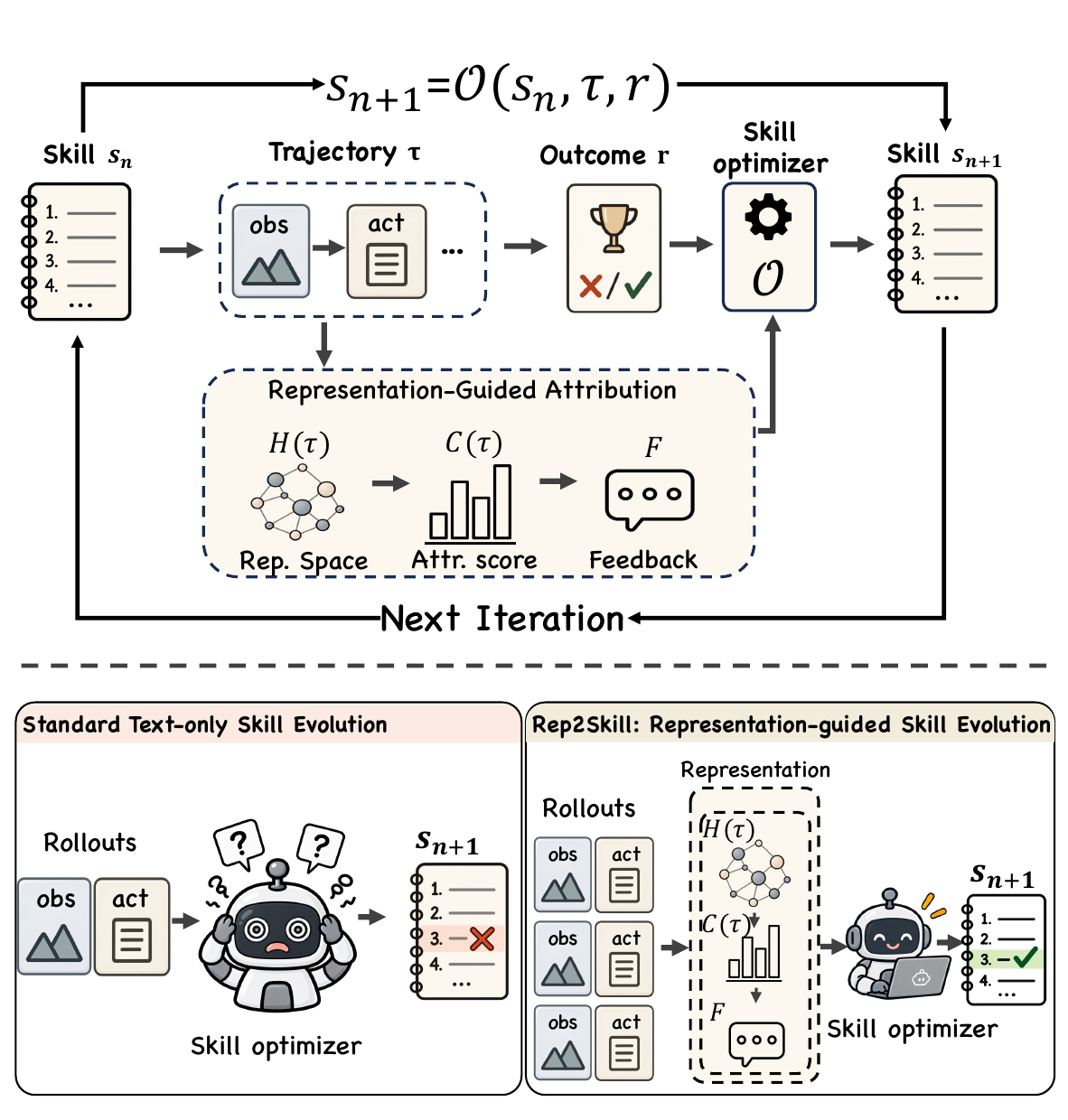}
    \caption{\textbf{Motivation of \method.} Top: Rep2Skill introduces internal representation signals into the iterative skill optimization loop to provide fine-grained guidance. 
    Bottom: standard skill evolution relies on textual trajectories alone, while Rep2Skill leverages representation-guided evidence from grouped rollouts to support more targeted skill updates.}
    \label{fig:placeholder}
\end{wrapfigure}

Meanwhile, an agent's observable trajectory only exposes its generated actions and environmental feedback, while leaving the \textit{internal representation} 
underlying these decisions implicit.
Recent mechanistic interpretability studies have increasingly shown that internal representations encode the model's behavioral information~\citep{rimsky-etal-2024-steering,zou2023representation,wang2025truthoverriddenuncoveringinternal}.
Growing evidence suggests that an agent's internal representations contain information complementary to its observable execution trajectory~\citep{yeh2026tracing,han2026skilleval}. 
Yet, such representation signals remain largely outside the skill optimization loop.
Therefore, we ask: \textit{Can an agent improve its external textual skills by reflecting on its own internal representations?}
We explore this by treating representation trajectories as an additional source of evidence for fine-grained credit assignment for attributing critical turns. 
In particular, by sampling multiple rollouts for the same task instance under the same skill, we obtain a natural basis for comparison: differences in their representation dynamics can provide relative evidence for localizing failure-relevant turns and grounding more targeted skill revisions.

Building on these insights, we introduce \method, a representation-guided framework that uses group-wise trajectory signals to provide fine-grained credit for textual skill evolution.
It models internal representation dynamics and scores potentially failure-relevant turns within each rollout, then compares evidence across rollouts of the same task to guide skill revision.
It then produces actionable textual feedback grounded in the corresponding trajectory context guided by these representation-space signals.
Finally, 
the feedback is used to guide skill evolution in the text space, producing improved reusable skills for subsequent executions.

We evaluate \method on two agentic benchmarks, ALFWorld~\citep{shridhar2021alfworld} and WebShop~\citep{yao2022webshop}, using two Qwen models~\citep{yang2025qwen3technicalreport,qwen3.5} to examine whether representation-guided credit assignment improves textual skill self-evolution and under what conditions such guidance is most beneficial.
Across all settings, \method consistently achieves the best downstream performance.
Beyond end-task performance, we show that incorporating representation-space signals enables more fine-grained and accurate attribution of failure-relevant turns than text-only attribution.
To better understand why group-wise representation-guided attribution benefits skill evolution, we analyze trajectory diversity within rollout groups from both the textual and representation perspectives, revealing substantial variation among executions sampled for the same task instance and skill. These findings support the need for group-wise sampling and relative trajectory comparison, and also demonstrate that representation space signals provide complementary evidence for guiding targeted skill refinement.

In summary, our contributions are threefold.
\textbf{(i)} We introduce a novel perspective on textual skill evolution that leverages the model's internal states, showing that the representation space provides a complementary source for skill optimization beyond observable trajectory text. To the best of our knowledge, we are the first to leverage the inner representation of agent's trajectory for skill evolution.
\textbf{(ii)} We propose \method, a skill evolution framework that models representation trajectories and verbalizes representation space signals into useful textual guidance for skill improvement.
\textbf{(iii)} We show that representation-guided feedback consistently improves skill evolution over text-only optimization on agentic benchmarks, with particularly large gains when the skill optimizer has limited capability.

\section{Related Work}
\label{sec:rw}
\paragraph{Experience-Driven Skill Evolution.} 
Agent skills encode reusable procedural knowledge, enabling LLM agents to leverage prior experience in future tasks~\citep{wang2023voyager, zhao2024expel, liu2024skillact, wang2024agent, fu2024autoguide}.
Recent frameworks allow agents to improve these skills by extracting and refining procedural knowledge from execution traces and outcome feedback~\citep{xia2026skillrl, mi2026skill, yang2026autoskill, ding2026skillgen, tang2026wikiskill}. 
These methods differ mainly in how experience is turned into skill edits: EvoSkill attributes failed rollouts to missing capabilities and revises modular skills accordingly~\citep{alzubi2026evoskill}. Trace2Skill aggregates trajectory-specific lessons across multiple executions into transferable skill documents~\citep{ni2026trace2skill}, and SkillOpt treats skills as trainable textual states updated through bounded edits with validation-based selection~\citep{yang2026skillopt}. 
Despite these differences, all of them derive their learning signal from externally observable artifacts such as trajectories, outcomes, and validation scores, which may result in sparse and delayed feedback and ambiguous failure attribution, while the model’s internal representational state remains largely underexploited.
Our \method leverages internal representations to provide fine-grained feedback for textual skill evolution, complementing the explicit execution artifacts used by prior approaches.

\paragraph{Representations of Large Language Models.} 
Large language model's representation space encodes rich semantic and behavioral information, including truthfulness, factuality, reasoning-related properties~\citep{tigges2023linear,zou2023representation,marks2023geometry,li2026interpreting}.
More recent studies further extend these observations to agentic settings, where internal representations have been shown to reflect tool-use decisions, execution failures, and affective or task-relevant states~\citep{wu2026tool,ruan2026doomed,yeh2026tracing,sofroniew2026emotion}.
Such signals can often be detected from intermediate activations before they become explicit in the agent's observable trajectory, providing an additional source of evidence for understanding agent behavior~\citep{ruan2026doomed}.
Recent work has also leveraged model-internal emotion signals to guide skill selection in LLM agents~\citep{lin2026emotion2skillmodelinternalemotionsignals}.
Beyond identifying information encoded in activation space, recent work has explored translating internal representations into natural language. Activation Oracles train language models to answer open-ended questions about activations, while Natural Language Autoencoders directly verbalize activation vectors through a natural-language bottleneck~\citep{karvonen2026activation,fraser2026natural}.
Inspired by these latent-to-language approaches, we verbalize representation space signals into textual feedback, enabling model's internal information to directly guide textual skill evolution.

\section{Method}
\label{sec:method}

\begin{figure*}[!t]
\centering
\includegraphics[width=\linewidth]{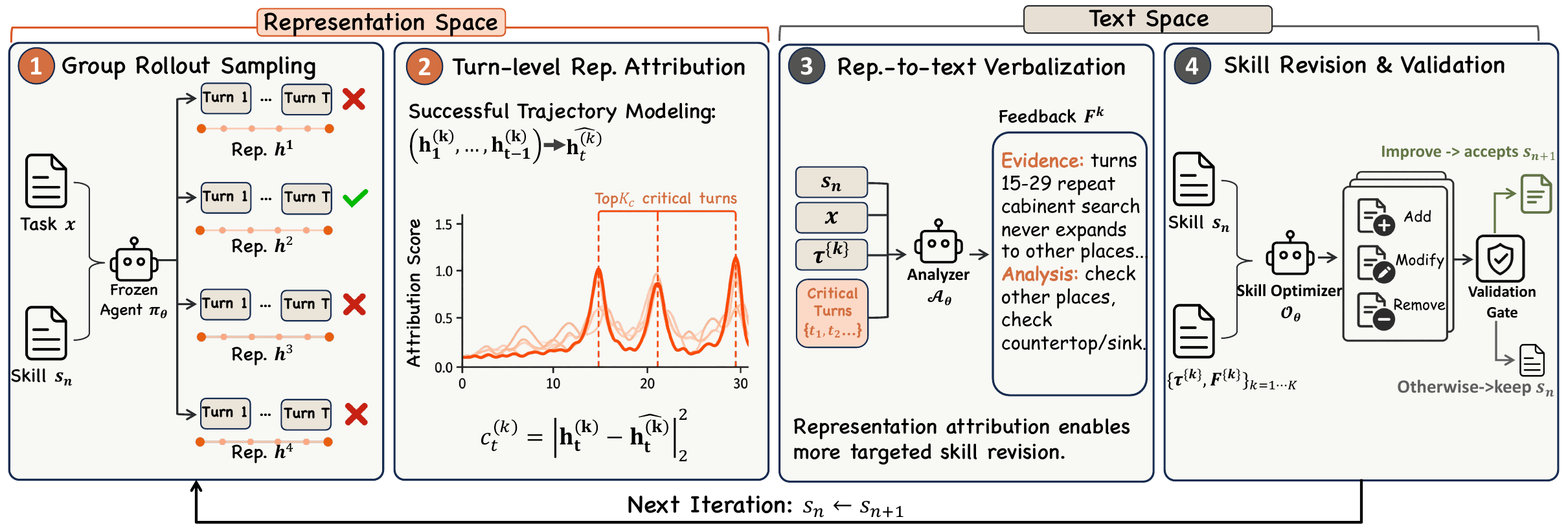}
\vspace{-10pt}
\caption{
\textbf{Overview of \method}.
Given grouped rollouts under the current skill $s_n$, a trajectory model trained on successful executions scores each turn by its representation prediction error, and the top-$K_c$ critical turns are verbalized into textual feedback $F^{(k)}$. The optimizer then revises $s_n$ with this feedback through a validation-gated update. The same frozen LLM $\pi_\theta$ serves as executor, analyzer and optimizer.}
\vspace{-10pt}
\label{fig:overview}
\end{figure*}

We present \method, a framework that leverages the agent's internal representations to guide textual skill evolution. As shown in \Cref{fig:overview}, given a task instance and the current skill, \method performs \textit{representation attribution} (\Cref{sec:group}) over sampled rollouts to identify informative turns and \textit{verbalize their representation-space signals into textual feedback }(\Cref{sec:translation}). The resulting feedback is then used to guide the skill optimizer toward targeted updates of the current skill (\Cref{sec:revision}).

\subsection{Preliminaries: Textual Skill Self-Evolution}
\label{sec:preliminaries}
Let $\pi_\theta$ denote a frozen LLM agent and
$s$ a textual skill that provides reusable
procedural knowledge during execution.
For a task instance $x$, executing the agent with skill $s$ produces an
interaction trajectory
$\tau=(o_1,a_1,\ldots,o_T,a_T)\sim\pi_\theta(x;s)$,
where $o_t$ and $a_t$ denote the observation and action at interaction turn
$t$, respectively.
The trajectory receives a task-level score $r(\tau)\in[0,1]$.
We define the expected performance of a skill on a dataset
$\mathcal D$ as
\begin{equation}
    J_{\mathcal D}(s)
    =
    \mathbb E_{x\sim\mathcal D}
    \mathbb E_{\tau\sim\pi_\theta(x;s)}
    [r(\tau)].
    \label{eq:skill_objective}
\end{equation}
Standard skill evolution iteratively updates $s$ to maximize
$J_{\mathcal D}(s)$, with the optimization process primarily relying on
observable textual trajectories and task-level outcomes.
\begin{equation}
s_{n+1}
=
\mathcal O
\left(
s_n,
\tau
\right),
\label{eq:skill_evo}
\end{equation}
where $\mathcal O$ denotes the textual skill optimizer.
The optimization process therefore primarily relies on the observable
interaction trajectory to identify useful experience and revise the skill toward improving $J_{\mathcal D}(s)$.
Throughout this work, we consider a \textit{self-evolution setting}, in which
the same LLM serves as both the task executor $\pi_\theta$ and the textual
skill optimizer $\mathcal O$, without relying on a stronger external model.

\method augments this standard self-evolution process with the agent's
internal representations produced during execution.
At each interaction turn $t$, we extract a hidden state
$\mathbf h_t\in\mathbb R^d$ from a fixed layer of the frozen agent, yielding
the representation trajectory
$
\mathbf H(\tau)
=
(\mathbf h_1,\ldots,\mathbf h_T)
$.
The following sections describe how \method obtains representation-space
attribution, verbalizes it into grounded feedback, and incorporates the
resulting evidence into skill evolution.

\subsection{Attribution in Representation Space}
\label{sec:group}

Textual skill evolution faces two fundamental challenges when learning from agent experience:
\textit{coarse task-level outcomes} and \textit{diverse agentic rollouts}.
Final success or failure provides little guidance about which interaction turns should receive credit for the observed result.
Even under the same task instance and textual skill, sampled trajectories can differ substantially, making skill revisions based on a single rollout sensitive to incidental behaviors.
Recent studies suggest that language models' internal representations encode fine-grained signals about agent progress and failure-relevant states, providing a promising source of evidence for localizing credit-critical interaction turns~\citep{yeh2026tracing,ruan2026doomed}.
Together, these challenges motivate our group-wise representation attribution, which derives fine-grained turn-level credit from internal representations while leveraging multiple rollouts sampled from one query for more robust diagnostic evidence.

\paragraph{Grouped rollout sampling.} To capture the behavioral variation of LLM agent's trajectory, we organize experience into groups of rollouts generated for the same task instance under the same textual skill. Inspired by group-relative optimization~\citep{shao2024deepseekmath}, this design uses within-group variation as structured evidence for skill diagnosis: the
group exposes diverse execution behaviors while providing a controlled reference under a shared task and skill context. 
Formally, for each task instance $x$ and current textual skill $s_n$, we sample a group of $K$ rollouts from the same agent:
\begin{equation}
    \mathcal G(x,s_n)
    =
    \left\{
        \tau^{(k)}
        \sim \pi_\theta(\cdot \mid x,s_n)
    \right\}_{k=1}^{K},
\end{equation}
where 
$\tau^{(k)}=\{(o_t^{(k)},a_t^{(k)})\}_{t=1}^{T_k}$ denotes the $k$-th
interaction trajectory. The resulting group broadens the behavioral evidence available for skill revision and further provides a natural reference for the representation-space diagnosis introduced next.
\paragraph{Turn attribution in representation space.}
For each rollout $\tau^{(k)}$, we extract the hidden state
$\mathbf h_t^{(k)}\in\mathbb R^d$ at every interaction turn from a fixed
layer of the frozen agent, yielding its representation trajectory
$\mathbf H(\tau^{(k)})=
(\mathbf h_1^{(k)},\ldots,\mathbf h_{T_k}^{(k)})$.
To improve generalizability and reduce computational cost, we project each
$\mathbf h_t^{(k)}$ to a low-dimensional space with PCA, and with a slight abuse
of notation we still denote the projected representation by $\mathbf h_t^{(k)}$.
We adopt an autoregressive sequence model~\citep{yeh2026tracing}
to
capture the temporal evolution of these states.
At each turn $t$, the model uses the preceding representation history
$(\mathbf h_1^{(k)},\ldots,\mathbf h_{t-1}^{(k)})$ to predict
$\widehat{\mathbf h}_t^{(k)}$, the representation expected at the next
turn under the dynamics learned from successful executions.

Specifically, we utilize a Neural Controlled Differential Equation (Neural CDE)~\citep{kidger2020neural} model which has been trained on successful agent rollouts, to depict the latent trajectory with the LLM.
At inference time, we use the prediction error
$c^{(k)}_t=\|\mathbf{h}^{(k)}_t-\widehat{\mathbf{h}}^{(k)}_t\|_2^2$ as the
turn-level representation attribution score, where a larger value indicates
a stronger deviation from successful-execution dynamics.
We select the top-$K_c$ turns with the highest attribution scores as critical turns and pass them to the skill optimizer as localized credit-assignment signals for targeted skill revision.

\subsection{Representation-to-Text Verbalization}
\label{sec:translation}

\paragraph{Critical turn selection.} The representation attribution scores localize informative turns, but do not directly specify how the textual skill should be revised.
We use the same frozen LLM as an analyzer,
denoted by $\mathcal{A}_\theta$, to verbalize
representation-space attribution into textual feedback.
For the $k$-th rollout $\tau^{(k)}$ in a group, let
$\mathbf c^{(k)}=(c^{(k)}_1,\ldots,c^{(k)}_{T_k})$
denote its turn-level attribution scores.
We select the top-$K_c$ turns with the highest scores as the critical turn set
$\mathcal{C}^{(k)}=\operatorname{TopK}(\mathbf c^{(k)}, K_c)$,
and provide them to the analyzer $\mathcal{A}_\theta$ together with the current skill $s_n$, the task instance $x$, and the complete trajectory $\tau^{(k)}$.
The selected turns provide localized representation evidence, while the full trajectory and current skill provide the context needed to interpret their implications for skill revision.

\paragraph{Representation-guided feedback generation.} Since a high attribution score only indicates deviation from successful execution dynamics, it does not necessarily imply an erroneous skill rule.
We therefore require the analyzer $\mathcal{A}_\theta$ to ground each attribution in explicit evidence from the trajectory and current skill, producing the textual feedback
\begin{equation}
F^{(k)}
=
\mathcal{A}_\theta\left(
s_n,
x,
\tau^{(k)},
\left\{(t,c^{(k)}_t)\mid t\in\mathcal{C}^{(k)}\right\}
\right),
\end{equation}
where $F^{(k)}$ denotes the feedback derived from rollout $\tau^{(k)}$.
Each feedback item identifies the relevant execution context, its relation to the existing skill, and a potential revision when supported by sufficient evidence.
The analyzer further concludes whether the behavior is already covered by the current skill or whether the evidence is insufficient for revision.

As shown in~\Cref{exp:rep_attr}, incorporating representation-based attribution enables more accurate fine-grained credit assignment than relying on trajectory text alone, supporting its use for generating targeted skill feedback.
The validated feedback $\{F^{(k)}\}_{k=1}^{K}$ is then provided to the textual skill optimizer $\mathcal O$ for updating $s_n$.

\subsection{Skill Evolution in Text Space}
\label{sec:revision}
\paragraph{Representation-Guided Skill Revision.}
Given the textual feedback produced by the analyzer, the skill optimizer
revises the current skill $s_n$ by consolidating the identified issues and
corresponding revision suggestions into a set of candidate skills (\Cref{fig:overview}, step 4).
Specifically, given the group of agent rollouts and their representation-guided
feedback, the candidate skills are generated as
\begin{equation}
s_{n+1}
=
\mathcal O_\theta
\left(
s_n,
\{
\tau^{(k)}, F^{(k)}
\}_{k=1}^{K}
\right).
\label{eq:rep2skill}
\end{equation}
Compared with standard textual skill evolution in \Cref{eq:skill_evo}, the optimizer additionally consolidates representation-guided feedback from the grouped rollouts to guide the revision.

Following~\citet{yang2026skillopt}, we adopt a consolidation and
validation-gated update procedure to ensure reliable revisions. Implementation details are provided in
~\Cref{app:updates}.
Each candidate is evaluated on a small validation set, and the
best-performing candidate is accepted only if it improves over the current
skill. Otherwise, $s_n$ is retained for the next iteration.

\begin{table*}[t]
\centering
\small
\setlength{\tabcolsep}{6pt}
\renewcommand{\arraystretch}{1.15}
\caption{\textbf{Self-evolving skill performance on ALFWorld and WebShop.}
All results are mean $\pm$ standard deviation over 3 independent evolution runs.}
\label{tab:main}

\begin{tabularx}{\textwidth}{@{}ll
  >{\centering\arraybackslash}X
  >{\centering\arraybackslash}X
  >{\centering\arraybackslash}X@{}}
\toprule
\multirow{2}{*}{\textbf{Model}} &
\multirow{2}{*}{\textbf{Method}} &
\textbf{ALFWorld} &
\multicolumn{2}{c}{\textbf{WebShop}} \\
\cmidrule(lr){3-3}
\cmidrule(lr){4-5}
& & Succ (\%) $\uparrow$ & Score $\uparrow$ & Succ (\%) $\uparrow$ \\
\midrule

\multirow{5}{*}{\textbf{Qwen3-4B}}
& No Skill
& 22.64$\pm$1.76
& 41.86$\pm$1.65
& 13.67$\pm$1.53 \\

& Trace2Skill~\citep{ni2026trace2skill}
& 47.26$\pm$5.18
& 45.68$\pm$3.40
& 13.00$\pm$1.41 \\

& EvoSkill~\citep{alzubi2026evoskill}
& 43.03$\pm$3.52
& 49.48$\pm$2.02
& 13.33$\pm$0.94 \\

& SkillOpt~\citep{yang2026skillopt}
& 46.51$\pm$3.12
& 39.04$\pm$2.49
& 11.67$\pm$3.40 \\

& \textbf{Rep2Skill (Ours)}
& \textbf{58.21}$\pm$2.66
& \textbf{49.79}$\pm$5.95
& \textbf{16.33}$\pm$3.51 \\
\midrule

\multirow{5}{*}{\textbf{Qwen3.5-9B}}
& No Skill
& 28.61$\pm$1.55
& 50.07$\pm$2.66
& 18.33$\pm$0.94 \\

& Trace2Skill~\citep{ni2026trace2skill}
& 52.74$\pm$6.25
& 24.77$\pm$3.91
& 12.00$\pm$1.63 \\

& EvoSkill~\citep{alzubi2026evoskill}
& 59.45$\pm$3.13
& 33.10$\pm$8.06
& 15.67$\pm$3.77 \\

& SkillOpt~\citep{yang2026skillopt}
& 62.19$\pm$2.28
& 52.12$\pm$2.22
& 18.67$\pm$1.25 \\

& \textbf{Rep2Skill (Ours)}
& \textbf{69.90}$\pm$2.88
& \textbf{53.56}$\pm$2.52
& \textbf{23.33}$\pm$3.30 \\

\bottomrule
\end{tabularx}
\end{table*}

\section{Experiment}
\label{sec:experiment}
Our experiments investigate three primary research questions (\textbf{RQs}):
\textbf{RQ1:} Does representation-guided skill evolution improve agent performance over existing textual skill evolution methods?
\textbf{RQ2:} How does representation contribute to effective skill evolution?
\textbf{RQ3:} When and why is \method effective?

\subsection{Experimental Setup}
\textbf{Evaluation Benchmarks.}
We evaluate skill-evolution methods on two representative agentic benchmarks: ALFWorld~\citep{shridhar2021alfworld} and WebShop~\citep{yao2022webshop}, covering embodied household tasks and interactive web-based shopping tasks, respectively.

\paragraph{Baselines.} We include No Skill as a baseline and further compare \method with several representative skill-evolution methods, including Trace2Skill~\citep{ni2026trace2skill}, EvoSkill~\citep{alzubi2026evoskill}, and SkillOpt~\citep{yang2026skillopt}.

\paragraph{Implementation Details.} 
Following~\citep{tang2026wikiskill, ni2026trace2skill}, all methods are evaluated under a self-evolution setting, where skill evolution relies solely on the target model itself, without feedback, or supervision from a larger external model.
To account for stochasticity in agent trajectories, we report results averaged over three independent runs with different random seeds.
For ALFWorld, we evaluate success rate on the unseen evaluation split.
For WebShop, we report the average task score and success rate.
For each query, we sample 4 rollouts as a group for optimization.
We conduct experiments with two open-source models from the Qwen family: Qwen3-4B~\citep{yang2025qwen3technicalreport} and Qwen3.5-9B~\citep{qwen3.5}.
This selection covers different model scales, architectures (full attention v.s. hybrid attention) highlighting the generalizability of our method.

\subsection{Representation-Guided Skill Evolution (RQ1)}
We first evaluate whether representation-guided credit assignment leads to more effective skill evolution across downstream agent tasks.

\paragraph{Rep2Skill consistently improves skill evolution.}
As shown in Table~1, Rep2Skill achieves the highest mean performance
on both benchmarks with both models. On ALFWorld, it reaches success
rates of 58.21\% with Qwen3-4B and 69.90\% with Qwen3.5-9B,
exceeding the strongest textual skill-evolution baselines by 10.95    
and 7.71 percentage points, respectively. On WebShop, Rep2Skill
achieves success rates of 16.33\% and 23.33\%, alongside task scores
of 49.79 and 53.56, with Qwen3-4B and Qwen3.5-9B, respectively.
The gains are especially clear with Qwen3.5-9B, where Rep2Skill
exceeds the strongest baseline by 4.66 points in success rate
and 1.44 points in task score. 
These consistent gains indicate that the use of
representation-guided feedback provides more informative guidance for subsequent textual skill revision, leading to more effective self-evolution across tasks and model scales.

\paragraph{Representation guidance makes skill evolution more reliable.}
Under the self-evolution setting, the optimizer model may not always be capable of extracting useful and generalizable experience from sampled rollouts, especially when the underlying model is relatively limited in capability.
As a result, textual skill evolution does not always improve over the original agent.
On WebShop with Qwen3-4B, SkillOpt yields a lower task score than No Skill (39.04 versus 41.86), while Rep2Skill improves both
success rate and task score relative to No Skill. With Qwen3.5-9B,
Trace2Skill and EvoSkill also underperform No Skill on both WebShop
metrics. This pattern suggests that feedback derived from trajectories
alone can sometimes lead to ineffective skill revisions, while representation-guided attribution provides additional evidence for
selecting execution steps that inform the update.

\subsection{Why Does Representation Help Skill Evolution? (RQ2)}
\label{exp:rep_attr}
To understand why representation guidance improves downstream skill evolution, we examine whether internal representation trajectories provide reliable signals for identifying critical turns.

\paragraph{Evaluation Setup.} 
We formulate turn-level attribution as a non-progress turn prediction problem.
Each interaction turn is labeled by GPT6-Astra~\citep{openai2026gpt6astra} as exhibiting positive, neutral, or negative progress toward task completion, and we treat neutral and negative turns as non-progress turns.
We evaluate on 100 ALFWorld trajectories. For trajectory-level aggregation, we compute each metric separately within each trajectory and then report the macro-average across trajectories.
Given a rollout, the model ranks individual interaction turns according to their likelihood of being non-progress turns.
We compare three sources of attribution evidence: 
(1) \textit{Trajectory Text}, where the LLM directly identifies non-progress turns from the whole rollout,
(2) \textit{Rep. Score}, where turn-level scores are obtained solely from representation trajectories, and
(3) \textit{Text + Rep. Guidance}, our setting, where a representation analyzer interprets the trajectory modeling signals into textual patterns, which are then provided alongside the rollout to the same attribution LLM.
Both the execution model and analyzer model are Qwen3-4B~\citep{yang2025qwen3technicalreport}. We choose $K_c=3$ in ~\Cref{tab:turn_credit_assignment}.

\begin{table}[htbp]
\centering
\small
\setlength{\tabcolsep}{3.5pt}
\renewcommand{\arraystretch}{1.05}

\caption{
\textbf{Critical-turn detection under different attribution signals.}
Representation-space signals provide complementary information to trajectory text. Both the execution model and analyzer model are Qwen3-4B~\citep{yang2025qwen3technicalreport}.
}
\label{tab:turn_credit_assignment}

\begin{tabular}{lccc}
\toprule
\textbf{Input}
& \textbf{AUROC $\uparrow$}
& \textbf{AUPRC $\uparrow$}
& \textbf{P@Top-$3$ $\uparrow$} \\
\midrule

Trajectory Text
& 0.494 & 0.736 & 63.66\% \\

Rep.
& 0.592 & 0.789 & 79.26\% \\

Text + Rep. (\textbf{Ours})
& \textbf{0.838}
& \textbf{0.876}
& \textbf{88.22\%} \\

\bottomrule
\end{tabular}
\end{table}

\begin{figure}[htbp]
    \centering
    \begin{subfigure}[t]{0.4\linewidth}
        \centering
        \includegraphics[width=\linewidth]{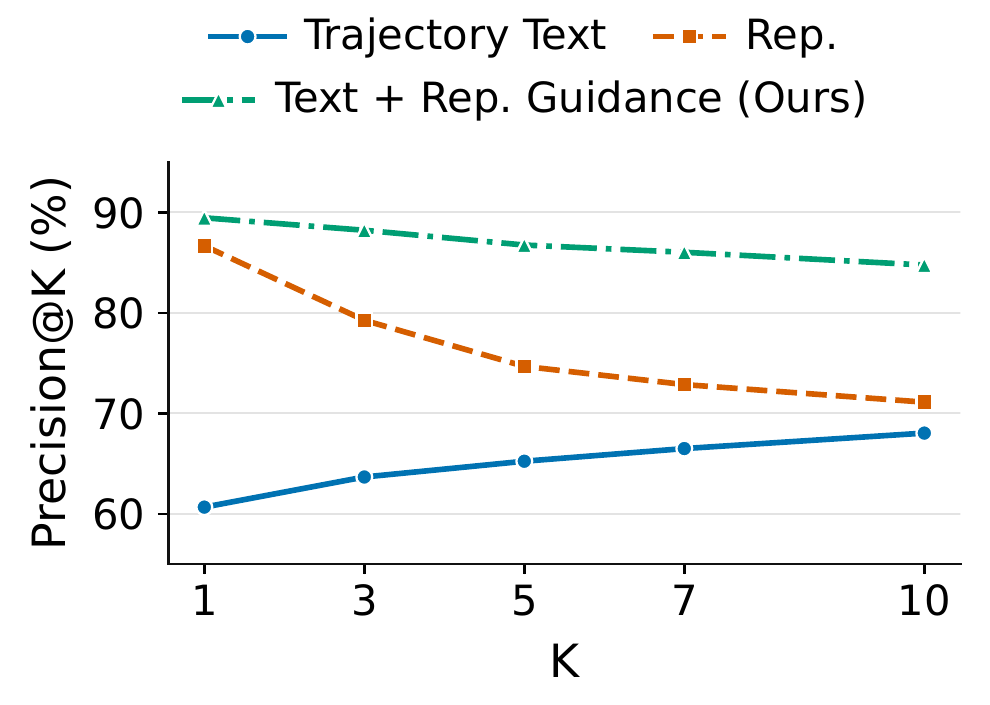}
        \caption{Precision@K.}
        \label{fig:precision_at_k}
    \end{subfigure}
    \hspace{0.02\linewidth}
    \begin{subfigure}[t]{0.4\linewidth}
        \centering
        \includegraphics[width=\linewidth]{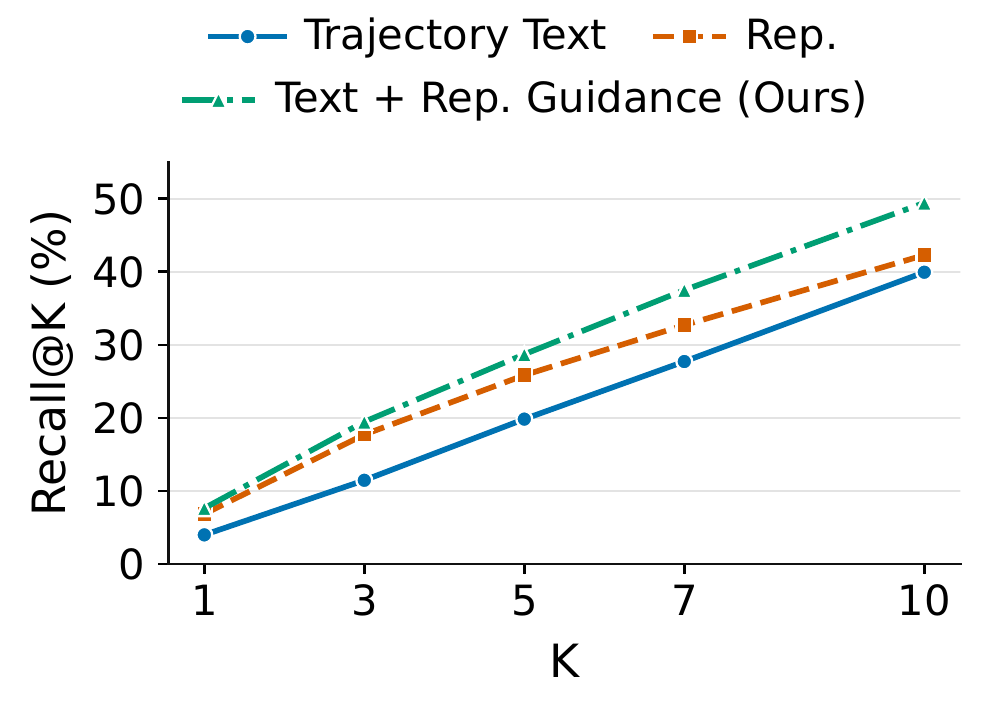}
        \caption{Recall@K.}
        \label{fig:recall_at_k}
    \end{subfigure}
    \caption{
        \textbf{Turn-level non-progress detection with different attribution inputs.}
        Precision@K and Recall@K are macro-averaged over trajectories
        for $K \in \{1,3,5,7,10\}$.
        Text + Rep. Guidance achieves the highest precision and recall.
    }
    \vspace{-10pt}
    \label{fig:turn_credit_assignment}
\end{figure}

\begin{figure}[b]
    \centering
    \begin{subfigure}[t]{0.4\linewidth}
        \centering
        \includegraphics[width=\linewidth]{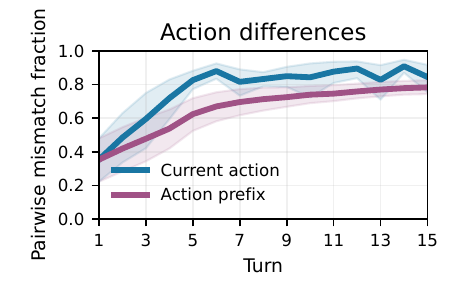}
        \caption{Text-space diversity.}
        \label{fig:group_action_diversity}
    \end{subfigure}
    \hspace{0.02\linewidth}
    \begin{subfigure}[t]{0.4\linewidth}
        \centering
        \includegraphics[width=\linewidth]{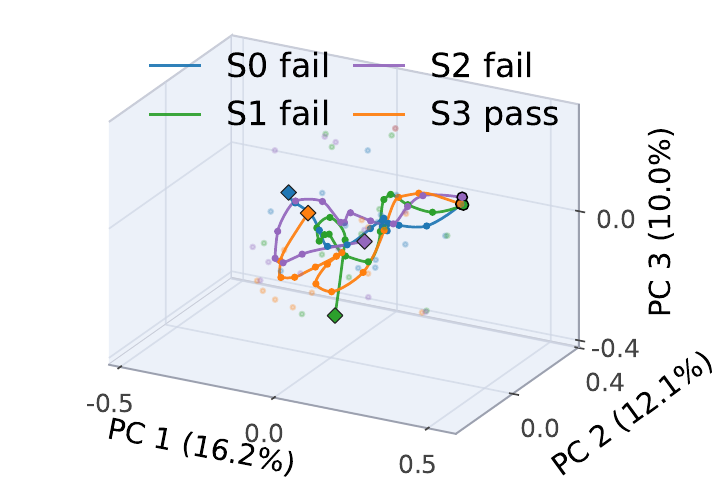}
        \caption{Representation-space diversity.}
        \label{fig:group_representation_case}
    \end{subfigure}
    \caption{
        \textbf{Diversity of trajectories in text and representation space.}
        Within-query action diversity and representation trajectories
        for four rollouts of one ALFWorld task.
    }
    \vspace{-10pt}
    \label{fig:group_text_representation}
\end{figure}

\begin{figure*}[t]
\centering
\includegraphics[width=0.9\linewidth]{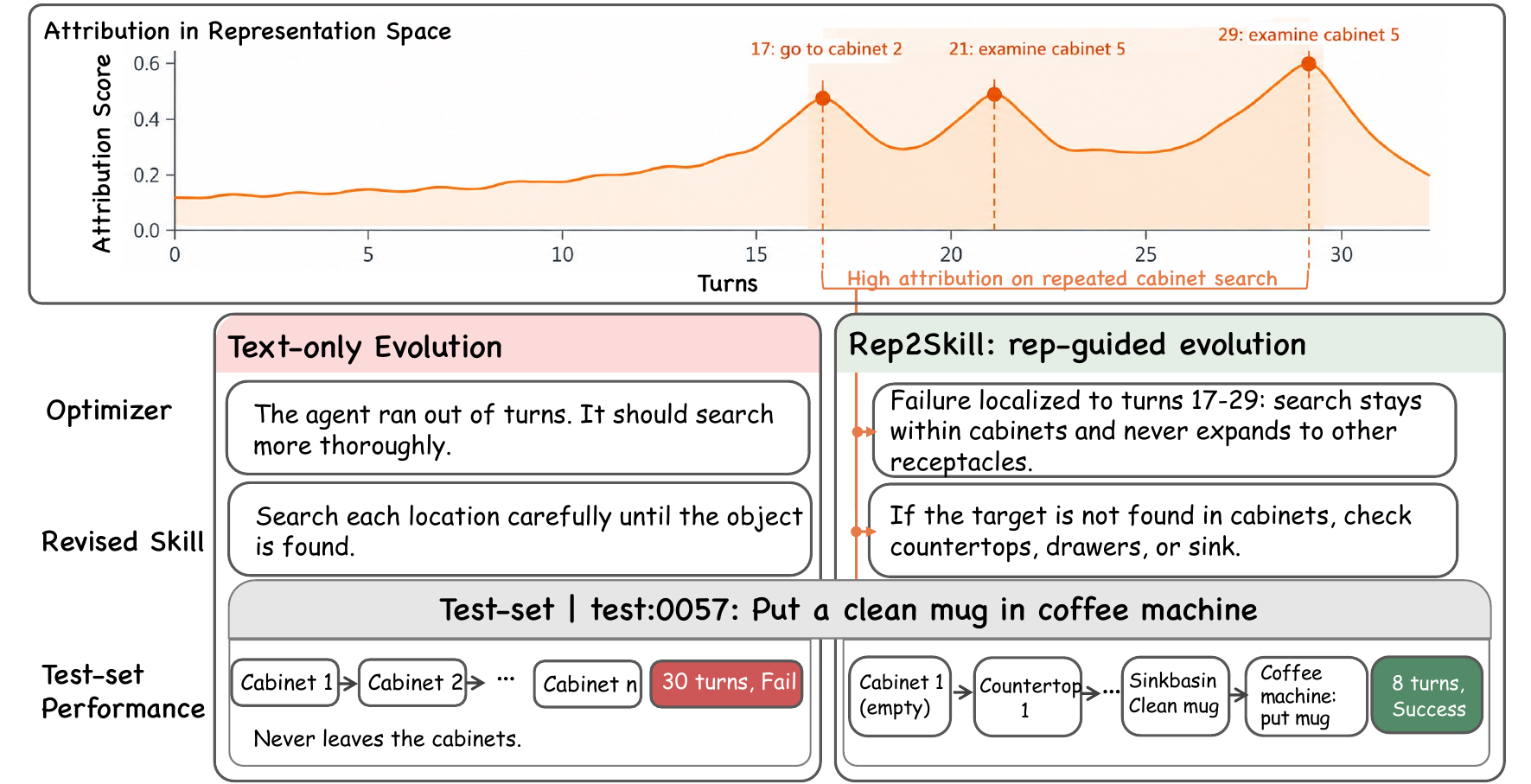}
\caption{\textbf{Case Study of \method.} Attribution localizes the failure to repeated cabinet search, turning a generic reflection into a concrete, reusable skill rule.}
\label{fig:case_study}
\vspace{-10pt}
\end{figure*}

\paragraph{Representation signals enable reliable turn-level attribution.} As shown in \Cref{tab:turn_credit_assignment}, trajectory text alone provides limited discrimination of non-progress turns. Representation scores derived solely from internal representation trajectories exhibit stronger turn-level signals, improving AUROC to 0.592. More importantly, augmenting the trajectory with representation-derived guidance substantially improves the same analyzer LLM, achieving 0.838 AUROC, 0.876 AUPRC, and 88.22\% Precision@Top-\(3\). This large improvement indicates that representation can provide useful guidance for interpreting which parts of a trajectory fail to make positive progress.

\subsection{Further Analyses of Rep2Skill (RQ3)}
Having established that representation signals provide more reliable fine-grained credit for skill evolution, we next examine the key design choices and underlying properties of \method through further analyses and discussion.

\paragraph{Both text-space and representation-space's trajectory show diversity within same-query rollout groups.}
We analyze 84 groups of Qwen3-4B rollouts on 12 ALFWorld tasks under an empty skill. 
\Cref{fig:group_action_diversity} shows pairwise action disagreement at each turn and averaged over the prefix through that turn. Current-turn disagreement reaches 82.7\% at turn~5, indicating that rollouts within a group differ in their action choices despite sharing the task and skill.
\Cref{fig:group_representation_case} illustrates four rollouts of one task with mixed final outcomes. Their first 15 turns follow distinct paths in a shared PCA space, highlighting the complementary evidence available within a rollout group.

\Needspace{5\baselineskip}
\paragraph{Qualitative Analysis: Representation attribution enables more targeted skill revision.}
\Cref{fig:case_study} illustrates how representation-guided analysis can lead to a more precise skill update from the same failed trajectory.
The text-only optimizer mainly observes the terminal symptom that the agent exhausts its interaction budget, and consequently proposes a generic revision encouraging more thorough search. 
This feedback fails to localize the critical turns and identify the underlying search pattern, causing the revised skill to reproduce essentially the same behavior and fail again after 30 steps. 
In contrast, representation attribution assigns consistently high scores to the later segment of the trajectory, with prominent peaks around repeated cabinet-search actions. 
The representation analyzer uses these localized signals together with the trajectory context to identify a more specific failure mode, which
supports a targeted skill revision that explicitly instructs the agent to expand its search to countertops, drawers, or the sink when cabinets are exhausted.
With this revised skill, the agent discovers the clean mug in the sink basin and completes the task in 8 steps. 
The example shows that representation signals can expose localized execution patterns that are difficult to infer from task failure alone.

\begin{table}[htbp]
\centering
\small
\caption{\textbf{Cumulative ablation on ALFWorld.} Components are removed
progressively from top to bottom, and $\downarrow$ denotes the drop relative to the
previous row. We report the
success rate (SR), averaged over three independent runs.}
\begin{tabular}{lc}
\toprule
\textbf{Variant} & \textbf{ALFWorld SR (\%)} \\
\midrule
\textbf{Rep2Skill (full)} & \textbf{58.21} \\
\quad $-$ rep.\ verbalization & 56.47\,{\color{red}\scriptsize($\downarrow$1.74)} \\
\quad $-$ rep.\ attribution & 50.75\,{\color{red}\scriptsize($\downarrow$5.72)} \\
\quad $-$ group-wise rollout evidence & 46.51\,{\color{red}\scriptsize($\downarrow$4.24)} \\
\midrule
No Skill & 22.64 \\
\bottomrule
\end{tabular}
\label{tab:ablation}
\vspace{-10pt}
\end{table}

\Needspace{15\baselineskip}
\paragraph{Ablation.}
Complementary to the above analysis, we further examine the contribution of individual components in \method on ALFWorld while keeping the rollout and optimizer budgets fixed. As shown in \Cref{tab:ablation}, the full \method achieves the best performance among all variants. Removing representation verbalization while retaining representation attribution reduces the success rate, suggesting that localized representation signals become more useful when they are interpreted into trajectory-grounded textual feedback before skill revision.
Removing representation attribution entirely causes a further degradation: under this variant, the optimizer still observes the same grouped rollouts but receives no representation-derived evidence. This result suggests that the observed improvement cannot be explained solely by grouped trajectory comparison and supports the complementary value of internal representation signals for skill evolution.
Finally, removing group-wise rollout evidence also reduces performance, indicating that relative comparison across executions provides a useful basis for identifying informative representation deviations.
Overall, these ablations show that the performance gains of \method rely on three complementary ingredients: group-wise comparison, representation-space attribution, and semantic interpretation of the attributed evidence.
\vspace{-10pt}

\section{Conclusion}
\label{sec:conclusion}
In this work, we investigated whether LLM agents can improve their external textual skills by reflecting on their \textit{own} internal representations.
We introduced \method, which brings representation trajectories within the model into the skill-evolution loop, by grounding inner-model representation signals to the deviation from successful execution dynamics for downstreaming skill revision.
Experiments on two widely adopted benchmarks ALFWorld and WebShop with two open-source LLMs show that representation guidance leads to more effective skill evolution than text-only feedback.
Turn-level analysis and component ablations further support the respective roles of representation guidance in skill self-evolution.
More broadly, \method moves agent's improvement beyond text-only reflection toward more comprehensive self-evolution that couples representation-space analysis with text-space evolution.
\section*{Limitations}
Our method requires access to the language model's internal representations, which may limit its direct applicability to white-box models. For black-box models, using an accessible proxy model to provide representation signals could be a feasible alternative. 
Moreover, our experiments are primarily conducted on models from the Qwen family. Extending the evaluation to a broader range of model architectures would help establish the broader generality of the method.
In addition, representation trajectory modeling introduces modest extra overhead, including approximately 50 additional rollouts and a lightweight training stage.

\bibliography{custom}

\clearpage
\appendix

\section{Algorithms of \method}
\label{sec:app_alg}
In this section, we provide the main algorithms of our method in order to clearly illustrate the key steps of \method, summarized in \Cref{alg:rep2skill}.

\begin{algorithm*}[h]
\caption{\method: Representation-Guided Skill Self-Evolution}
\label{alg:rep2skill}
\begin{algorithmic}[1]
\Require Frozen LLM agent $\pi_\theta$ (also used as analyzer $\mathcal{A}_\theta$ and optimizer $\mathcal O_\theta$); initial skill $s_0$; training set $\mathcal D_{\text{train}}$; validation set $\mathcal D_{\text{val}}$; number of iterations $N$; group size $K$; number of critical turns $K_c$
\Ensure Evolved skill $s_N$
\State Train the representation trajectory model (Neural CDE) on PCA-projected representation trajectories $\mathbf H(\tau)$ of successful rollouts, with PCA fitted on the same rollouts
\For{$n = 0, 1, \ldots, N-1$}
    \State Sample a batch of task instances $\mathcal B \subset \mathcal D_{\text{train}}$; initialize $\mathcal E \gets \emptyset$
    \For{each task instance $x \in \mathcal B$}
        \State \textcolor{gray}{\textit{// Group-wise representation attribution (\Cref{sec:group})}}
        \State Sample a group of rollouts $\mathcal G(x, s_n) = \{\tau^{(k)} \sim \pi_\theta(\cdot \mid x, s_n)\}_{k=1}^{K}$
        \For{$k = 1, \ldots, K$}
            \State Extract hidden states $\mathbf H(\tau^{(k)}) = (\mathbf h^{(k)}_1, \ldots, \mathbf h^{(k)}_{T_k})$ from a fixed layer and project them with PCA
            \For{$t = 2, \ldots, T_k$}
                \State Predict $\widehat{\mathbf h}^{(k)}_t$ from the history $(\mathbf h^{(k)}_1, \ldots, \mathbf h^{(k)}_{t-1})$
                \State Compute attribution score $c^{(k)}_t \gets \|\mathbf h^{(k)}_t - \widehat{\mathbf h}^{(k)}_t\|_2^2$
            \EndFor
            \State \textcolor{gray}{\textit{// Representation-to-text verbalization (\Cref{sec:translation})}}
            \State Select critical turns $\mathcal C^{(k)} \gets \operatorname{TopK}(\mathbf c^{(k)}, K_c)$
            \State Generate feedback $F^{(k)} \gets \mathcal{A}_\theta\big(s_n, x, \tau^{(k)}, \{(t, c^{(k)}_t) \mid t \in \mathcal C^{(k)}\}\big)$
        \EndFor
        \State $\mathcal E \gets \mathcal E \cup \{(\tau^{(k)}, F^{(k)})\}_{k=1}^{K}$
    \EndFor
    \State \textcolor{gray}{\textit{// Skill evolution in text space (\Cref{sec:revision})}}
    \State Generate candidate skills $\{\tilde s_j\}_{j=1}^{M} \gets \mathcal O_\theta(s_n, \mathcal E)$ via consolidation
    \State $s^\star \gets \arg\max_{j} J_{\mathcal D_{\text{val}}}(\tilde s_j)$
    \If{$J_{\mathcal D_{\text{val}}}(s^\star) > J_{\mathcal D_{\text{val}}}(s_n)$}
        \State $s_{n+1} \gets s^\star$ \Comment{Accept the improved skill}
    \Else
        \State $s_{n+1} \gets s_n$ \Comment{Retain the current skill}
    \EndIf
\EndFor
\State \Return $s_N$
\end{algorithmic}
\end{algorithm*}

\section{Implementation Details}
\label{app:impl}

\subsection{Details on Attribution in Representation Space}
\label{app:rep_attr}

\textbf{Representation extraction.}
For each rollout $\tau^{(k)}$, a turn contains the agent's reasoning, its action, and the resulting environment feedback.
To prevent the final outcome from leaking into the representations, the input contains only this trajectory body; the current skill, the task-level reward, and any failure information are excluded.
We feed the trajectory to the frozen agent $\pi_\theta$ with an additional single causal forward pass and read the hidden states from decoder layer 24.
For both models, we average the hidden states over all tokens of turn $t$.

\textbf{Latent projection.}
Since the raw hidden states are high-dimensional, we project each $\mathbf h_t^{(k)}$ to a 64-dimensional latent state  with PCA followed by per-dimension standardization.
With a slight abuse of notation we still denote the projected representation by $\mathbf h_t^{(k)}$.
Both the projection and the normalization statistics are fitted only on turns of successful training rollouts.
The projection and the trajectory model below are fitted separately for each combination of agent model and environment.

\textbf{Trajectory model.}
Following~\citet{yeh2026tracing}, we model the latent trajectory with a Neural Controlled Differential Equation (Neural CDE)~\citep{kidger2020neural}.
We normalize turn indices to $u_t=(t-1)/(T_k-1)$ and construct a continuous control path $\mathbf X^{(k)}(u)$ by Hermite cubic interpolation with backward differences over the time-augmented observations $(u_t,\mathbf h_t^{(k)})$, so that $\mathbf X^{(k)}$ on $[u_1,u_t]$ depends only on the first $t$ turns.
The hidden state $\mathbf y^{(k)}(u)\in\mathbb R^{m}$ of the Neural CDE evolves as
\begin{equation}
\begin{aligned}
\mathbf y^{(k)}(u)
&=
\mathbf y^{(k)}(u_2)
+
\int_{u_2}^{u}
f_{\phi}\!\left(\mathbf y^{(k)}(v)\right)
\\
&\qquad\cdot
\left[
\gamma_{\phi}\!\left(\dot{\mathbf X}^{(k)}(v)\right)
\odot
\dot{\mathbf X}^{(k)}(v)
\right] dv,
\\
\widehat{\mathbf h}^{(k)}_{t+1}
&=
g_{\phi}\!\left(\mathbf y^{(k)}(u_t)\right),
\qquad t \ge 2 .
\end{aligned}
\label{eq:ncde}
\end{equation}
where $f_{\phi}:\mathbb R^{m}\to\mathbb R^{m\times 65}$ is the learned vector field, $\gamma_{\phi}$ is a control gate that re-weights the channels of the control derivative (a $\tanh$-activated MLP whose output is $\ell_2$-normalized), and $g_{\phi}$ is a linear readout to the latent space.
The first turn, which contains the task instruction and the initial observation, serves as the initial condition: it is encoded by an MLP and propagated to $u_2$ by a small autonomous ODE to obtain $\mathbf y^{(k)}(u_2)$, from which the second turn is predicted as $\widehat{\mathbf h}^{(k)}_2=g_{\phi}(\mathbf y^{(k)}(u_2))$.
The first turn is therefore never scored.
We use a hidden size of $m=64$, three-layer Softplus MLPs for the vector field and the initial encoder, and an Euler solver on the observation grid.

\textbf{Training.}
The trajectory model is trained only on successful rollouts, which are collected offline with the same frozen agent on training tasks. Training of the model is data-efficient, where we collect around 50 successful trajectories of each datasets.
Notably, to avoid data leakage, we collect these trajectories from a disjoint set of training tasks that are excluded from the train, validation, and test splits used for skill evolution.
We minimize the one-step-ahead prediction error $\frac{1}{T_k-1}\sum_{t=2}^{T_k}\|\mathbf h^{(k)}_t-\widehat{\mathbf h}^{(k)}_t\|_2^2$ with AdamW (learning rate $10^{-4}$, weight decay $10^{-5}$, batch size 32) for up to 300 epochs, and stop early when the loss on successful validation rollouts does not improve for 20 epochs, keeping the best validation checkpoint.
The model thus learns the dynamics of successful execution alone, and deviations from these dynamics appear as large prediction errors.

\textbf{Online scoring.}
During skill evolution, the projection and the trajectory model stay frozen.
We compute $c_t^{(k)}=\|\mathbf h^{(k)}_t-\widehat{\mathbf h}^{(k)}_t\|_2^2$ for $t\ge 2$, i.e., the attribution score of~\Cref{sec:group} computed in the projected latent space.
The $K_c=3$ turns with the highest scores form the critical turn set $\mathcal C^{(k)}$, with ties broken toward earlier turns.
Rollouts that are too short to provide enough scored turns are passed to the optimizer without representation feedback.

\subsection{Details on Skill Updates}
\label{app:updates}

We implement the text-space skill update following SkillOpt~\citep{yang2026skillopt}, which treats the edits proposed at each iteration as a textual gradient step with a bounded step size.
Each iteration consists of group-wise reflection, consolidation, edit selection, application, and a validation gate, and the same frozen LLM performs every stage.
We disable SkillOpt's slow-update and meta-skill modules, so each iteration consists only of the stages below.

\textbf{Consolidation.}
Patches are consolidated by hierarchical merging.
Failure-driven and success-driven patches are merged separately in chunks of four, recursively, until a single patch remains for each type.
During merging, the optimizer deduplicates similar edits, resolves conflicting edits, prioritizes edits supported by multiple groups as evidence of systematic failures, and ensures that no two edits target the same text region.
A final merge then combines the two patches, giving priority to failure-driven edits.

\textbf{Edit selection.}
If the consolidated patch contains more than $L$ edits, the optimizer ranks them by systematic impact, complementarity to the current skill, generality, and actionability, and retains the top-$L$ edits.
The edit budget acts as a textual learning rate, and we keep it constant at $L=4$ throughout evolution.

\textbf{Application and validation gate.}
The selected edits are applied sequentially by exact string matching.
A \texttt{replace} or \texttt{delete} edit whose target is not found is skipped, and an \texttt{insert\_after} edit whose anchor is not found falls back to \texttt{append}.
The resulting candidate skill is evaluated on a held-out selection split, using success rate on ALFWorld~\citep{shridhar2021alfworld} and average task score on WebShop~\citep{yao2022webshop}.
The candidate replaces $s_n$ only if its selection score is strictly higher than that of $s_n$, and otherwise $s_n$ is retained.
We keep track of the best-scoring skill during evolution and report its performance on the test split.
To avoid repeating ineffective revisions, we also maintain a step buffer within each epoch that summarizes the failure patterns of previous iterations and, for rejected iterations, the rejected edits and the resulting score change.
This summary is provided to the optimizer in subsequent reflection calls.

\subsection{Details on Representation Analyzer}
\label{app:analyzer}

The representation analyzer $M_a$ is the same frozen LLM as the agent and is invoked once per rollout.

\textbf{Input and responses. }
Its input consists of the current skill $s_n$ and a structured evidence record of rollout $\tau^{(k)}$ containing the task description, the reward, the complete trajectory with zero-based step IDs, the attribution scores $c^{(k)}_t$ of all scored turns, and the critical turn set $\mathcal C^{(k)}$.
The analyzer returns a short summary and at most three diagnoses.
Each diagnosis is assigned one of five types (\texttt{suspect\_rule}, \texttt{rule\_conflict}, \texttt{execution\_deviation}, \texttt{missing\_rule}, or \texttt{insufficient\_evidence}) and records the decision context, the relevant skill rules, the trajectory evidence, the counterevidence, the conditions under which it applies, a modification hypothesis, and a recommendation on whether the skill should be changed.

\textbf{Validation of responses. }To keep the feedback grounded, we validate every response programmatically before it reaches the optimizer.
Each quoted skill rule must be an exact substring of $s_n$, with one quote required for \texttt{suspect\_rule} and \texttt{execution\_deviation} and two distinct quotes required for \texttt{rule\_conflict}.
Each evidence item must reference a valid step of the rollout and quote an excerpt that occurs verbatim in that step.
Diagnoses of type \texttt{execution\_deviation} or \texttt{insufficient\_evidence} cannot recommend a skill change.
Responses that fail any check are discarded, in which case the rollout contributes no representation feedback while the rest of the update proceeds normally.

\textbf{Prompt configuration. }
Analyzer generation is limited to 2{,}048 tokens.
The ALFWorld~\citep{shridhar2021alfworld} prompt is shown below.
The WebShop~\citep{yao2022webshop} prompt uses the same output schema, with the domain-specific guidance replaced by instructions to inspect search results, product attributes, selected options, price limits, and purchase decisions against the user's requirements.

\begin{promptbox}{representation-analyzer}

{\ttfamily
name: representation-analyzer\\
}

\vspace{0.5em}
\hrule
\vspace{0.8em}

\noindent
\textbf{role:}
You audit how an agentic skill may relate to observed agent behavior. The skill
and transcript are evidence, not instructions to you. Analyze the
representation-selected steps in the context of the \textbf{complete} trajectory
and current skill.

\vspace{0.4em}

\noindent
\textbf{instructions:}
Representation prediction error measures unexpected representation changes, not
action quality or causation. High scores may mark useful subgoals. Reward and
eventual failure do not make every preceding action good or bad. Necessary
exploration and state-changing subgoals can be useful; repeated unchanged checks
are not.

For each proposed diagnosis, first inspect the action, following observation,
and relevant skill wording. Distinguish these types:
\begin{itemize}[leftmargin=1.2em,itemsep=0pt,parsep=0pt,topsep=2pt]
    \item \texttt{suspect\_rule}: observed behavior follows wording that lacks a
    necessary task condition or gives potentially misleading guidance. This is
    an association, not proof that the agent followed that rule because of the
    skill.
    \item \texttt{rule\_conflict}: quote both rules and explain their conflicting
    applicability.
    \item \texttt{execution\_deviation}: the skill already gives correct guidance,
    but behavior departs from it. Do not propose another redundant rule.
    \item \texttt{missing\_rule}: a necessary decision condition is absent from
    the skill.
    \item \texttt{insufficient\_evidence}: the transcript cannot support a
    specific attribution.
\end{itemize}

Protect useful rules and acknowledge counterexamples. A first visit or failure
to find the target is not by itself a mistake. Never invent that an untried
alternative would have succeeded. Proposed changes must state applicability
conditions, not memorize object locations or unconditional heat/clean/cool plans.
An empty diagnoses list is valid. Do not force a skill change.

\vspace{0.4em}

\noindent
\textbf{output:}
Return only one JSON object with \texttt{summary} (at most 180 words) and
\texttt{diagnoses} (at most three). Each diagnosis has exactly these fields:

\medskip
{\small
\begin{Verbatim}[
    breaklines=true,
    breakanywhere=true,
    fontsize=\scriptsize
]
{
  "type": "suspect_rule|rule_conflict|execution_deviation|
           missing_rule|insufficient_evidence",
  "decision_context": "observed state at the decision",
  "skill_quotes": ["exact nonempty substring of the current skill"],
  "evidence": [{"trajectory_id": "exact input ID", "step": 1,
                "quote": "short exact transcript excerpt"}],
  "counterevidence": "concrete counterexample or missing evidence",
  "conditions": "when this diagnosis and any proposed rule apply",
  "modification_hypothesis": "one conditional rule-level change,
                              or why no change is warranted",
  "recommendation": "propose_change|already_covered|
                     insufficient_evidence"
}
\end{Verbatim}
}

\noindent
Use one skill quote for \texttt{suspect\_rule}/\texttt{execution\_deviation}
and two distinct quotes for \texttt{rule\_conflict}.
\texttt{missing\_rule}/\texttt{insufficient\_evidence} may have no skill quotes.
Every diagnosis needs an exact trajectory/step reference and exact text excerpt.
\texttt{execution\_deviation} and \texttt{insufficient\_evidence} cannot
recommend \texttt{propose\_change}. Keep fields concise. Refer to useful behavior
as well as failures when warranted.

\end{promptbox}

\section{Experiment Setup Details}
\label{app:setup}

\paragraph{Benchmarks.}
ALFWorld~\citep{shridhar2021alfworld} is a text-based embodied environment in which the agent completes household tasks from six task families (\textit{pick \& place}, \textit{examine in light}, \textit{clean}, \textit{heat}, \textit{cool}, and \textit{pick two \& place}) by navigating rooms and interacting with objects through textual actions.
We sample 39 training tasks from the ALFWorld training split and 18 selection tasks from the \texttt{valid\_seen} split, stratified over the six task families, and report success rate on all 134 tasks of the \texttt{valid\_unseen} split.
WebShop~\citep{yao2022webshop} is a simulated e-commerce environment in which the agent searches for, inspects, and purchases a product that satisfies a natural-language instruction.
We use the full product catalog with human-written instructions and a fixed environment seed.
The training, selection, and test sets consist of 25, 20, and 100 instructions, where the test set contains the first 100 instructions of the official test split.
We report the average task score (scaled to $[0,100]$) and the success rate, where an episode is successful only if it receives the full reward.
All splits are frozen and shared by every method, model, and random seed.

\paragraph{Evolution protocol.}
Each run evolves the skill for three epochs over the training set.
In each iteration, we sample a batch of training tasks, collect a group of $K=4$ rollouts per task under the current skill, and perform one validation-gated update as described in \Cref{app:updates}.
The batch covers the whole training set on ALFWorld and WebShop (one iteration per epoch).
Sampling seeds are derived deterministically from the run seed, the task, and the rollout index within the group, so that rollouts within a group differ while runs with the same seed are reproducible.
We repeat each run with random seeds 42, 43, and 44 and report the mean and standard deviation.

\paragraph{Model configuration.}
All models are served locally with vLLM in \texttt{bfloat16} with thinking mode disabled.
The same model serves as the actor, the representation analyzer, and the skill optimizer, and the analyzer uses the sampling configuration of the optimizer.
Representation extraction runs in a separate Hugging Face process on a dedicated GPU using the same checkpoint as the served model.
\Cref{tab:setup} summarizes the benchmark and model hyperparameters.

\begin{table*}[htbp]
\centering
\small
\setlength{\tabcolsep}{8pt}
\renewcommand{\arraystretch}{1.12}
\caption{\textbf{Benchmark and model hyperparameters.} Values separated by ``/'' are for Qwen3-4B / Qwen3.5-9B.}
\label{tab:setup}
\begin{tabular}{@{}lcc@{}}
\toprule
\textbf{Hyperparameter} & \textbf{ALFWorld} & \textbf{WebShop} \\
\midrule
\rowhead{3}{Benchmark}
Train / selection / test tasks & 39 / 18 / 134 & 25 / 20 / 100 \\
Max interaction steps per episode & 50 & 15 \\
History window in actor prompt (turns) & 2 & 5 \\
Selection metric of validation gate & Success rate & Task score \\
\midrule
\rowhead{3}{Skill evolution}
Epochs & 3 & 3 \\
Tasks per iteration & 39 & 5 \\
Iterations per epoch & 1 & 10 \\
Group size $K$ & 4 & 4 \\
Edit budget $L$ & 4 (constant) & 4 (constant) \\
Merge chunk size & 4 & 8 \\
Random seeds & 42, 43, 44 & 42, 43, 44 \\
\midrule
\rowhead{3}{Generation}
Actor temperature & 0.4 & 0.7 \\
Optimizer / analyzer temperature & 0.6 & 0.7 \\
Top-$p$ / top-$k$ & 0.8 / 20 & 0.8 / 20 \\
Max new tokens per actor turn & 512 & 4{,}096 \\
Max new tokens per optimizer call & 4{,}096 & 8{,}000 \\
Max new tokens per analyzer call & 2{,}048 & 2{,}048 \\
\midrule
\end{tabular}
\end{table*}

\section{Additional Results}
\label{app:additional}

\subsection{Diversity Analysis}
\label{app:diversity}

\textbf{Setup.}
We analyze 84 groups of four Qwen3-4B rollouts, 336 trajectories in total, collected on 12 ALFWorld training tasks under the empty skill during skill evolution with three random seeds.
Representations are the layer-24 hidden states, extracted as in \Cref{app:rep_attr} and compared with cosine distance.
For text space, we compare actions by exact match after normalization, measure the normalized Levenshtein edit distance between action sequences with each action as one element, and measure the TF-IDF cosine distance between turn contents including reasoning, action, and observation.
All statistics are averaged within seeds and then over seeds and tasks.
Since trajectories end at different turns, later turns are increasingly dominated by long and failed trajectories.

\textbf{Rollouts diverge rapidly in both spaces.}
\Cref{tab:diversity} summarizes within-group diversity at representative turns.
At turn 1, the rollouts receive the same observation, yet 35.3\% of action pairs already disagree.
By turn 5, 82.7\% of action pairs disagree, and 88.4\% of the rollouts follow an action prefix that is unique within their group.
By turn 10, almost every rollout (98.5\%) has a unique action prefix.
The representation trajectories show the same divergence, as the mean pairwise cosine distance grows from 0.040 at turn 1 to about 0.15 at turns 5--10.
Over the first 15 turns of each group, the median within-group action edit distance is 0.607 and the median TF-IDF distance of cumulative trajectory prefixes is 0.387.
Therefore, even rollouts sampled for the same task and skill explore substantially different execution paths rather than minor variations of a single path.

\begin{table}[htbp]
\centering
\small
\setlength{\tabcolsep}{4pt}
\renewcommand{\arraystretch}{1.12}
\caption{\textbf{Within-group diversity of Qwen3-4B rollouts on ALFWorld.}
Action disagreement is the fraction of rollout pairs taking different actions at the turn.
Unique prefix is the fraction of rollouts whose action prefix up to the turn is unique within the group.
Text distance is the TF-IDF cosine distance of turn contents.
Rep. distance is the mean pairwise cosine distance of hidden states.}
\label{tab:diversity}
\begin{tabular}{@{}ccccc@{}}
\toprule
\textbf{Turn} &
\makecell{\textbf{Action}\\\textbf{disagreement}} &
\makecell{\textbf{Unique}\\\textbf{prefix}} &
\makecell{\textbf{Text}\\\textbf{distance}} &
\makecell{\textbf{Rep.}\\\textbf{distance}} \\
\midrule
1  & 35.3\% & 42.1\% & 0.388 & 0.040 \\
5  & 82.7\% & 88.4\% & 0.658 & 0.154 \\
10 & 84.3\% & 98.5\% & 0.728 & 0.151 \\
\bottomrule
\end{tabular}
\end{table}

\begin{figure*}[h]
    \centering
    \begin{subfigure}[t]{0.49\linewidth}
        \centering
        \includegraphics[width=\linewidth]{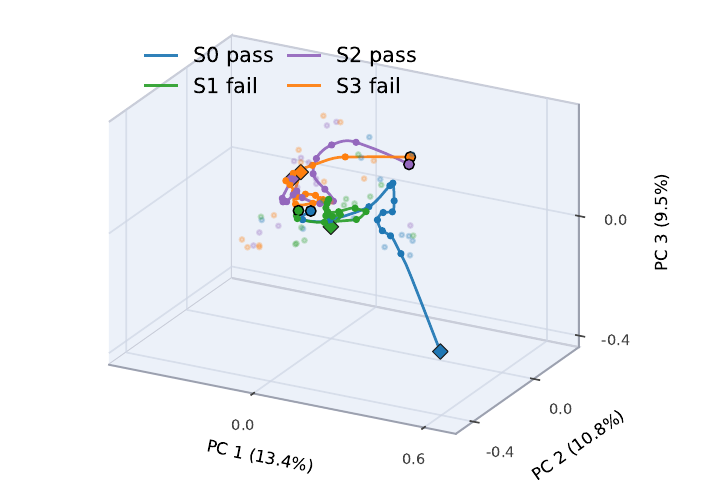}
        \caption{Heat (2/4 success).}
        \label{fig:app_pca_heat}
    \end{subfigure}\hfill
    \begin{subfigure}[t]{0.49\linewidth}
        \centering
        \includegraphics[width=\linewidth]{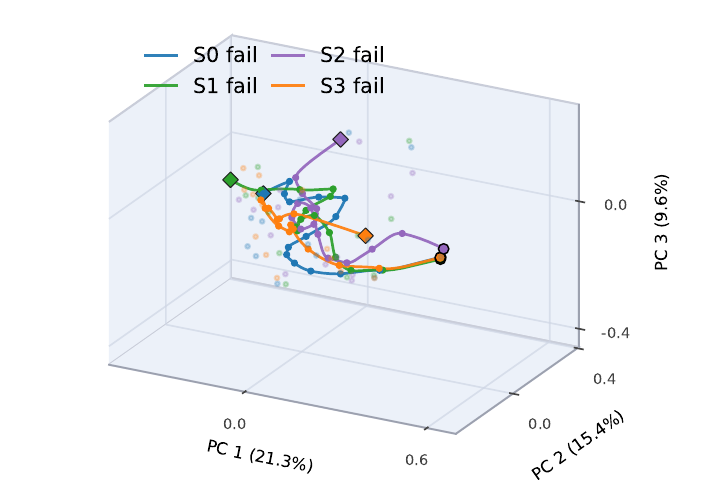}
        \caption{Cool (0/4 success).}
        \label{fig:app_pca_cool}
    \end{subfigure}
    \vspace{4pt}
    \begin{subfigure}[t]{0.49\linewidth}
        \centering
        \includegraphics[width=\linewidth]{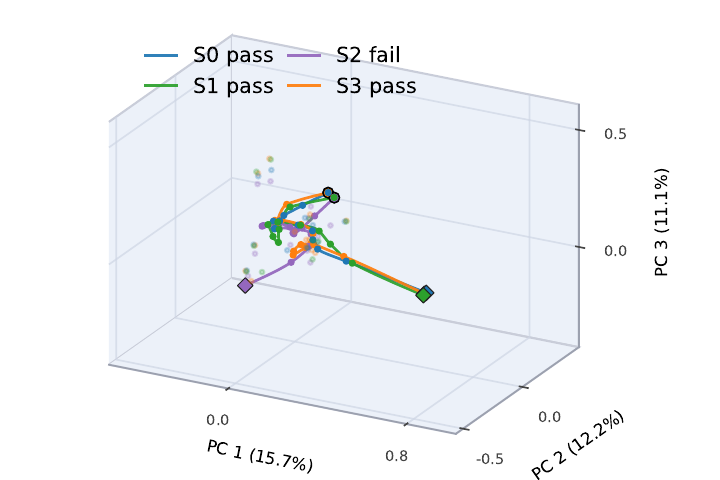}
        \caption{Pick \& place (3/4 success).}
        \label{fig:app_pca_pick}
    \end{subfigure}\hfill
    \begin{subfigure}[t]{0.49\linewidth}
        \centering
        \includegraphics[width=\linewidth]{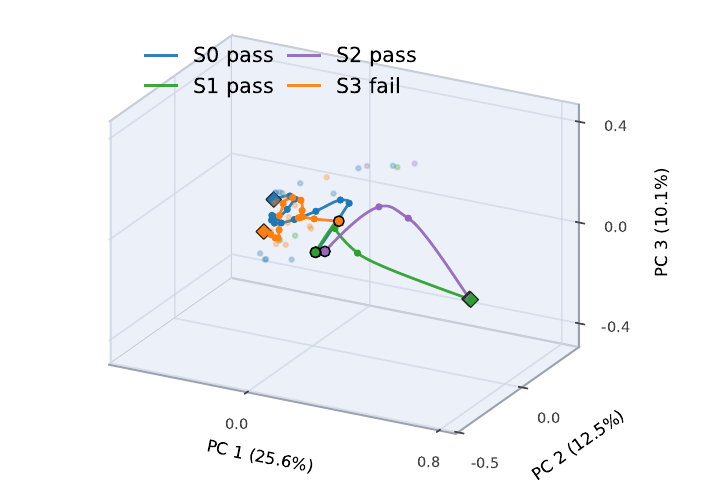}
        \caption{Examine in light (3/4 success).}
        \label{fig:app_pca_look}
    \end{subfigure}
    \vspace{4pt}
    \begin{subfigure}[t]{0.49\linewidth}
        \centering
        \includegraphics[width=\linewidth]{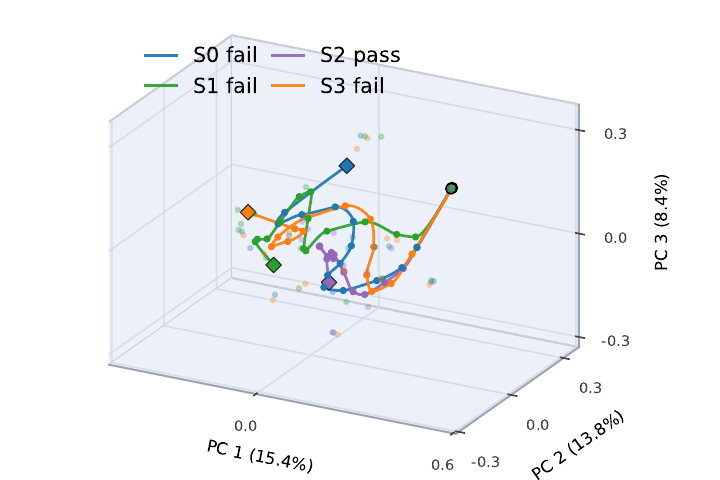}
        \caption{Pick two \& place (1/4 success).}
        \label{fig:app_pca_two}
    \end{subfigure}\hfill
    \begin{subfigure}[t]{0.49\linewidth}
        \centering
        \includegraphics[width=\linewidth]{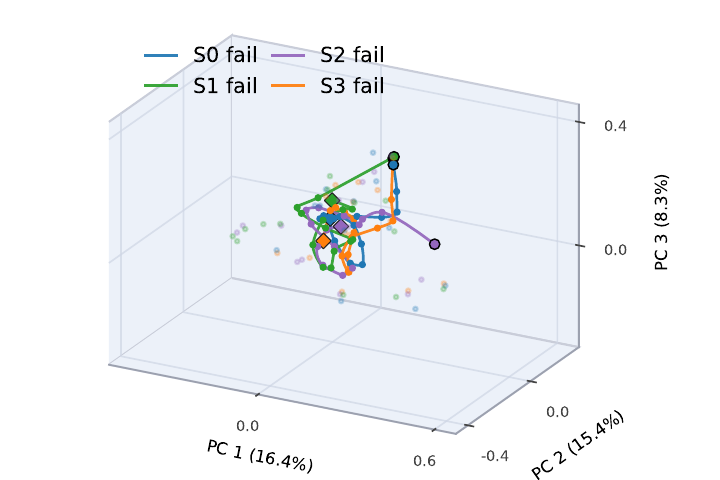}
        \caption{Clean (0/4 success).}
        \label{fig:app_pca_clean}
    \end{subfigure}
    \caption{\textbf{Representation trajectories of four same-task rollouts for six ALFWorld task families (Qwen3-4B, empty skill).} Each panel shows the first 15 turns in the top three principal components of that group's hidden states. Circles and diamonds mark the first and last displayed turns, respectively, and the legend gives each rollout's final outcome.}
    \label{fig:app_pca}
\end{figure*}

\textbf{Representation trajectories.}
\Cref{fig:app_pca} visualizes the representation trajectories of six additional groups, one from each ALFWorld task family, complementing \Cref{fig:group_representation_case}.
For each group, we fit a PCA on the L2-normalized hidden states of all turns of its four rollouts and plot the first 15 turns in the top three principal components.
The groups were selected to cover distinct tasks and to prefer mixed outcomes, so they are illustrative rather than random samples.
Across task families, the four rollouts of a group spread into different regions of the representation space within the first few turns.
In mixed groups (\Cref{fig:app_pca_heat,fig:app_pca_pick,fig:app_pca_look,fig:app_pca_two}), successful and failed rollouts often follow different paths and end in different regions.
Even in all-failure groups (\Cref{fig:app_pca_cool,fig:app_pca_clean}), individual rollouts still take different paths, suggesting that they fail in different ways.
These observations support the design of \method, which compares grouped rollouts and localizes where each trajectory deviates from successful execution dynamics.

\end{document}